\documentclass[sn-mathphys,Numbered]{sn-jnl}

\usepackage{graphicx}%
\usepackage{multirow}%
\usepackage{amsmath,amssymb,amsfonts}%
\usepackage{amsthm}%
\usepackage{mathrsfs}%
\usepackage[title]{appendix}%
\usepackage{xcolor}%
\usepackage{textcomp}%
\usepackage{manyfoot}%
\usepackage{booktabs}%
\usepackage{algorithm}%
\usepackage{algorithmicx}%
\usepackage{algpseudocode}%
\usepackage{listings}%
\usepackage{newtxtext,newtxmath}
\usepackage[colorinlistoftodos]{todonotes}%
\usepackage{caption}
\usepackage{subcaption}
\usepackage{lineno}

\theoremstyle{thmstyleone}%
\newtheorem{theorem}{Theorem}
\newtheorem{proposition}[theorem]{Proposition}%

\theoremstyle{thmstyletwo}%

\theoremstyle{thmstylethree}%
\newtheorem{definition}{Definition}%

\begin{document}

\title[Article Title]{A Graph Neural Network-Based QUBO-Formulated Hamiltonian-Inspired Loss Function for Combinatorial Optimization using Reinforcement Learning}


\author[1]{\fnm{Redwan Ahmed} \sur{Rizvee}}\email{rizvee@cse.du.ac.bd}

\author[1]{\fnm{Raheeb} \sur{Hassan}}\email{raheeb-2017915005@cs.du.ac.bd}

\author[1]{\fnm{Md. Mosaddek} \sur{Khan}}\email{mosaddek@du.ac.bd}

\affil*[1]{\orgdiv{Department of Computer Science and Engineering}, \orgname{University of Dhaka}, \orgaddress{\city{Dhaka}, \country{Bangladesh}}}








\abstract{Quadratic Unconstrained Binary Optimization (QUBO) is a generic technique to model various NP-hard Combinatorial Optimization problems (CO) in the form of binary variables. \textcolor{black}{Ising Hamiltonian is used to model the energy function of a system}. \textcolor{black}{QUBO to Ising Hamiltonian is regarded as a technique to solve various canonical optimization problems through quantum optimization algorithms}. Recently, PI-GNN, a generic framework, has been proposed to address CO problems over graphs based on Graph Neural Network (GNN) architecture. \textcolor{black}{They introduced} a generic QUBO-formulated Hamiltonian-inspired loss function that was \textcolor{black}{directly optimized} using GNN. \textcolor{black}{PI-GNN is highly scalable but there lies a noticeable decrease in the number of satisfied constraints when compared to problem-specific algorithms and becomes more pronounced with increased graph densities.} \textcolor{black}{Here, We identify a behavioral pattern related to it and devise strategies to improve its performance.} \textcolor{black}{Another group of literature uses Reinforcement learning (RL) to solve the aforementioned NP-hard problems using problem-specific reward functions}. \textcolor{black}{In this work, we also focus on creating a bridge between the RL-based solutions and the QUBO-formulated Hamiltonian.} We formulate and empirically evaluate the compatibility of the QUBO-formulated Hamiltonian as the generic reward function in the RL-based paradigm in the form of rewards. Furthermore, we also introduce a novel Monty Carlo Tree Search-based strategy with GNN where we apply a guided search through manual perturbation of node labels during training. We empirically evaluated our methods and observed up to 44\% improvement in the number of constraint violations compared to the PI-GNN. }

\keywords{Hamiltonian Function, Deep Reinforcement Learning, Graph Neural Network, Monte Carlo Tree Search}



\maketitle

\section{Introduction}



Combinatorial Optimization (CO) is indeed a crucial field in mathematics and computer science with a wide range of applications in solving real-life problems. At its core, CO deals with finding the best possible solution from a finite set of possibilities, where the goal is to optimize some objective function subject to a set of constraints. One of the key challenges in CO is that as the problem size grows, the number of possible solutions increases exponentially, making it computationally infeasible to explore all possibilities. Therefore, CO seeks efficient algorithms and techniques to find near-optimal solutions.

One powerful framework for addressing CO problems is Quadratic Unconstrained Binary Optimization (QUBO), which transforms complex combinatorial problems into a quadratic equation involving binary variables. QUBO has become the standard format for optimization using quantum computers, such as the quantum approximate optimization algorithm (QAOA) and quantum annealing (QA) \cite{gabor2022approximate}. It is also applicable to gate arrays and digital annealers \cite{verma2021constraint}. QUBO allows for the formulation of hard combinatorial optimization problems as polynomial-sized linear programs, providing a generalized framework for solving these problems \cite{diaby2016advances}. \textcolor{black}{QUBO and Ising Hamiltonian have a very close relationship in modeling a diverse set of optimization problems and are often used interchangeably or together \cite{schuetz2022combinatorial}. Ising Hamiltonian focuses on modeling the energy stability of a system in the form of identifying the spins ($\{-1, 1\}$) of the qubits. We have already mentioned that the terms QUBO and Hamiltonian are closely related. After converting a QUBO objective or loss function into an Ising Hamiltonian, various quantum optimization algorithms are used to solve it \cite{speziali2021solving}. Due to this close connection, the relevant problems are addressed as problem Hamiltonian and the objective functions are regarded as Hamiltonian Objective or Cost function \cite{schuetz2022combinatorial, schuetz2023reply}. Research has shown that quantum solution systems struggle to scale with the number of variables and to maintain precision in the solutions they provide \cite{schuetz2023reply}. Thus, recent works have been addressing different strategies to model QUBO-based Hamiltonians \cite{schuetz2022combinatorial}.}

In general terms, the goal of QUBO is to find a diverse set of solutions that meet a target metric or goal, making it suitable for high-level decision-making. QUBO has found applications in various domains, including logistics, finance, and artificial intelligence. In logistics, it aids in optimizing supply chain routes, minimizing transportation costs, and improving inventory management. In finance, QUBO assists in portfolio optimization, risk management, and trading strategies. \cite{verma2021efficient} In AI, it plays a pivotal role in solving complex problems such as feature selection, clustering, and tuning of the parameters of the machine learning models \cite{zaman2021pyqubo}.

Graph Neural Networks (GNNs) are a class of deep learning models designed for processing and analyzing structured data represented as graphs. They have gained significant popularity for their ability to capture complex relationships and dependencies in graph-structured data, making them highly versatile for a wide range of applications. In particular, GNNs excel in tasks involving graph data, such as social network analysis, recommendation systems, and, as discussed here, combinatorial optimization \cite{li2023survey, xu2023preference, gao2023sruh}. When it comes to addressing combinatorial optimization problems, \textcolor{black}{\cite{schuetz2022combinatorial} has introduced a pioneering approach (PI-GNN) harnessing the power of Graph Neural Networks (GNNs) to address the usage of Hamiltonian cost function during training}. In their paper, they showcase how GNNs can effectively tackle quadratic unconstrained binary optimization (QUBO) and polynomial unconstrained binary optimization (PUBO) problems. The key innovation lies in their utilization of a relaxation strategy applied to the problem Hamiltonian, resulting in a differentiable loss function that serves as the foundation for GNN training. The paper's impressive results demonstrate that their GNN-based optimizer not only matches the performance of existing solvers but often outperforms them, particularly in scenarios such as maximum cut and maximum independent set problems. What truly sets this approach apart is its marked scalability, allowing it to effectively tackle combinatorial optimization problems involving millions of variables. This work represents a significant advancement, merging deep learning techniques with concepts from statistical physics to address NP-hard problems and offering promising possibilities for the field of combinatorial optimization.

In a recent critical review paper by Boettcher et al. (2023) \cite{boettcher2023inability}, it was demonstrated that PI-GNN, performs less effectively than traditional greedy algorithms when solving the Max-Cut problem. Angelini et al. \cite{angelini2023modern} raised similar concerns about GNN-based solutions, particularly their relative inefficiency compared to classical greedy algorithms, as discussed in the context of the Maximum Independent Set problem. Both pieces of literature highlight the notion that in very narrowly defined problem scenarios, such as Max-Cut and Maximum Independent Set, traditional greedy algorithms may outperform PI-GNN. However, it is noteworthy that the authors of PI-GNN have responded to these critiques in their subsequent work \cite{schuetz2023reply}, arguing that focusing solely on the performance in these specific, curated scenarios overlooks the broader generality and scalability of their proposed framework. To bolster their perspective, they have presented empirical results that illustrate the advantages of PI-GNN in more general and scalable problem settings, suggesting that a comprehensive evaluation of the framework is necessary to appreciate its full merit.

On the other hand, another widely explored approach to solving optimization problems is Reinforcement Learning (RL). RL aims to learn policies that maximize expected rewards, which can often be derived from the objective function of the optimization problem \cite{kotary2021end}. In this setup, \textcolor{black}{an agent's} states represent current solutions, actions correspond to decisions, and rewards measure solution quality. RL algorithms like Q-learning or deep reinforcement learning (DRL) methods guide the agent to explore and exploit solutions efficiently.  Bello et al. (2016) extended the pointer network architecture to create an actor-critic RL framework for training an approximate Traveling Salesman Problem (TSP) solver, using tour length as a reward signal \cite{bello2016neural}. Subsequent work \cite{kool2018attention} introduced improvements in accuracy using a graph attention network for two-dimensional Euclidean TSP, approaching optimal solutions for problems up to 100 nodes. Additionally, a multi-level RL framework was used to analyze TSP variants with hard constraints. In terms of a general solving framework, Drori et al. \cite{drori2020learning} proposed a technique that uses a Graph Attention Network (GAT) base encoder architecture to encode and generate the node feature vectors, upon which they apply an attention-based decoding mechanism to greedily select the nodes and apply the node labelings. The approach however requires reward functions to be specifically tailored for each problem and thus hampers generality.


Taking into account the previous discussion, it is obvious that trivial GNN based methods such PI-GNN, although they are widely applicable and versatile, often struggle with accuracy. Meanwhile, RL-based methods, despite their high accuracy and scalability, are not as widely applicable. In this \textcolor{black}{paper}, we try to bridge this gap by addressing the combinatorial optimization problem by extending some of the aforementioned methodologies. In particular, we suggest three distinct improvements. Firstly, we identify a crucial flaw in the early stopping strategy of PI-GNN which causes it to underperform in denser graphs. We propose a fuzzy early-stopping strategy as an alternative to the fixed tolerance value strategy. Secondly, being inspired by the formulation stated in \cite{drori2020learning} and \cite{khalil2017learning}, we propose a modified generic framework that works with QUBO-formulated Hamiltonian as the generic reward function. This offers more accuracy compared to the previous approach at the cost of higher run time. Finally, we propose a Monty Carlo Tree Search-based strategy with GNN through manual perturbation to avoid local minimas. This performs best in terms of violating the least constraints but also comes at the expense of higher run time. \textcolor{black}{Within this context, it is worth noting that the quality of the number of satisfied constraints can also be quite important for various applications in both online and offline settings. For example, In optimal sensor placement problems, it is more important to have better sensor assignments than faster reporting of the results \cite{speziali2021solving}. Sensor placement problem is a constraint-dependent version of the vertex cover problem which can be formulated in the QUBO-based Hamiltonian.} For empirical evaluation, we evaluated our approaches considering Max-Cut as the benchmarking problem and have witnessed up to $44\%$ improvement over PI-GNN in reducing the number of violated constraints. It is noteworthy to mention that, similar to PI-GNN, all the presented proposals are generic in manner and can be extended to a wide group of graph-based canonical optimization problems that can be formulated in QUBO.

\section{Background}
This section lays the groundwork for our discussion by taking a quick look at the fundamentals of combinatorial optimization, graph neural networks and reinforcement learning. Furthermore, we also discuss the methodologies used by Schuetz et al. (2023) \cite{schuetz2022combinatorial} and Drori et al. (2020) \cite{drori2020learning} to solve the combinatorial optimization problem.

\subsection{Combinatorial Optimization}
Combinatorial Optimization seeks to find the optimal solution from a finite set of possibilities while adhering to specific constraints. Common combinatorial optimization problems include the maximum cut problem (Max-Cut), the maximum independent set problem (MIS), the minimum vertex set cover problem, the maximum clique problem, the set cover problem, the traveling salesman problem (TSP), etc. Most of these problems include settings where a set of decisions have to be made and each set of decisions yields a corresponding cost or profit value, that is to be maximized or minimized. As the problem size grows, the size of the set of possible decision variables increases, and so does the set of feasible solutions, exponentially in fact. This makes exhaustive exploration impractical. Moreover, for most of these problems, finding an optimal solution in polynomial time is known to be NP-complete. This motivates the development of various approximation algorithms.

\subsection{Quadratic Unconstrained Binary Optimization (QUBO)}
Although many approximate algorithms exist for solving combinatorial problems, most of them are tailored to the specific problems they try to solve with limited generalizability \cite{ghaffari2016improved,karakostas2009better}. That is where Quadratic Unconstrained Binary Optimization (QUBO) comes in, being able to transform a large class of such NP-complete combinatorial problems into quadratic equations involving binary variables. If we take $X = (x_1, x_2, ...)$ to be a vector binary decision variables, we can represent the cost function of a QUBO problem with the Hamiltonian \textcolor{black}{expressed in Equation \ref{eqn:qubo_hamiltonian}},
\begin{equation} \label{eqn:qubo_hamiltonian}
    H_{\text{QUBO}} = X^T Q X = \sum_{i=1}^{n} \sum_{j=1}^{n} x_i Q_{ij} x_j,
\end{equation}
where \(Q_{ij}\) represents the coefficients of the quadratic terms which encodes the constraints of the problem we are trying to solve. Then the problem becomes effectively reduced to finding $X$ which minimizes $H_{\text{QUBO}}$. The $Q$-matrix can be generated specifically for problems as per requirement. Moreover, extra constraints may be included in the form of penalty terms in the objective function as is often needed in real world combinatorial optimization problems.

\subsection{Graph Neural Networks (GNNs)}
In recent years, Graph Neural Networks (GNNs) have emerged as a powerful framework for analyzing structured data, particularly in the context of graph-structured data. Graphs are mathematical structures that consist of nodes (representing entities) and edges (representing relationships or connections between entities). GNNs have gained significant attention due to their ability to model and extract valuable information from such complex, interconnected data.

On a high level, they are a family of deep learning models capable of learning representations from graph structured data. While a single layer GNN is able to encapsulate a node's features in a one-hop neighbourhood, typically multi-layered stacked GNNs are used to capture information from a larger neighbourhood of the node. Formally, in a graph \(G = (V, E)\), at layer $k = 0$, each \textcolor{black}{node} $v \in V$ is represented by some initial representation $h_v^0$, usually derived from the node's label or given input features. GNNs then iteratively update each node's representation by some parametric function \(f_{\theta}^k\) using,
\begin{equation}
    h_v^k = f_{\theta}^k(h_v^{k-1}, \{h_u^{k-1}| u \in \mathcal{N}(v)\}),
\end{equation}
for layers $k = 1, 2, ..., K$ where $\mathcal{N}(v)$ denotes the neighbours of node $v$. Finally, the last ($K^{\text{th}}$) layer's output is used in a problem-specific loss function, and stochastic gradient descent is used to train the weight parameters. For example, for classification problems, the soft-max function combined with cross-entropy loss is often used \cite{mao2023cross}.

\subsection{GNNs for Combinatorial Optimization}
\label{sec:pignn}
Combining the power of GNNs with QUBO, Schuetz et al. introduced PI-GNN to solve various combinatorial optimization problems \cite{schuetz2022combinatorial}. It leverages a relaxation strategy applied to the problem Hamiltonian, resulting in a differentiable loss function for GNN training. Formally, a loss function $L_{\text{QUBO}}(\theta)$ is defined based on $H_{\text{QUBO}}$ as,
\begin{equation}
\label{eqn:pi_gnn_loss}
    H_{\text{QUBO}} \rightarrow L_{\text{QUBO}}(\theta) = \sum_{i, j} p_i(\theta) Q_{ij} p_j(\theta),
\end{equation}
where $p_i$ is the node embeddings generated by the final layer of the GNN ($p_i = h_i^K \in [0, 1]$). Now, $L_{\text{QUBO}}(\theta)$ is differentiable with respect to the parameters of the GNN model $\theta$. After unsupervised training, a simple projection step ($x_i = int(p_i)$) is used to enforce integer variables. As for the GNN architecture, PI-GNN uses graph convolutional networks (GCN) with the following update step,
\begin{equation}
    h_v^k = \sigma (W_k \sum_{u \in \mathcal{N}(v)} \frac{h_u^{k-1}}{|\mathcal{N}(v)|} + B_k h_v^{k-1}),
\end{equation}
where $W_k$ and $B_k$ are trainable weight matrices, $\sigma(.)$ is some non-linear activation function such as sigmoid or ReLU for middle layers and SoftMax for the final layer. For the actual gradient descent process, ADAM is used. PI-GNN excels in complex combinatorial optimization problems, such as maximum cut and maximum independent set problems, and is characterized by its scalability.


A significant issue associated with Equation \ref{eqn:pi_gnn_loss} is the omission of the actual node projection strategy from the loss function, denoted as $L_{QUBO}$. Therefore, the projection strategy employed to label the node set upon completion of training will dictate the quality of the solution to the variable assignment problem.

\subsubsection{Reinforcement Learning (RL)}

Reinforcement Learning (RL) is a computational approach where an agent learns to make sequential decisions by interacting with an environment. At the core of RL are \textit{states}, representing the circumstances, \textit{actions}, the choices available to the agent, and \textit{rewards}, the immediate feedback based on these actions. The agent's objective is to maximize cumulative rewards over time. Through trial and error, RL algorithms, such as Q-learning \cite{watkins1992q} or Deep Q Networks \cite{mnih2013playing}, enable the agent to learn optimal strategies by exploring the environment, experiencing different states, and refining its decision-making policy. The process involves constant decision-making based on received rewards, learning from these experiences, and adjusting strategies accordingly.

\subsubsection{RL for Combinatorial Optimization}
\label{sec:rlco}
In 2020, Drori et al. proposed a generic RL-based framework (that we address as GRL) to solve a wide variety of combinatorial optimization problems \cite{drori2020learning}. The key insight was that if we turn the combinatorial optimization problem into a node selection problem on a graph, we can use RL to learn how to choose nodes to get the optimal result. Here the \textit{state} is the set of nodes already chosen, the \textit{action} being choosing a node to add to this set and the \textit{reward} is some function of these set of nodes specific to the problem at hand that ensures that maximizing it ensures that we get the optimal solution. A Graph Attention Network (GAT) \cite{velickovic2017graph} (a type of GNN) was employed for encoding to generate node feature vectors for greedily choosing nodes at each step. Each layer $k=1, ... , K$ of GNN updates the feature vector of the $v$-th node as,
\begin{equation}
    h_v^{k} = \alpha_{vv} \Theta^k h_v^{k-1} + \sum_{u \in \mathcal{N}(v)} \alpha_{vu} \Theta^k h_u^{k-1}
\end{equation}
where $\Theta^k$ is a learnable weight matrix, and $\alpha_{vu}^k$ are the attention coefficients, defined as,
\begin{equation}
    \alpha_{vu}^k = \frac{\text{exp}(\sigma({z^k}^T \left[\Theta^k h_v^k, \Theta^k h_u^k \right]))}{\sum_{w \in \mathcal{N}(v)} \text{exp}(\sigma({z^k}^T \left[\Theta^k h_v^k, \Theta^k h_w^k \right]))}.
\end{equation}
The decoder on the other hand uses a simple attention mechanism, where we compute the attention coefficients as follows,
\begin{equation}
\label{eqn:grl_decode}
    \alpha_{vu}^{\text{dec}} = C \text{ tanh} \left( \frac{(\Theta_1 h_v)^T(\Theta_2 h_u)}{\sqrt{d_h}} \right)
\end{equation}
where $\Theta_1, \Theta_2 \in \mathbb{R}^{d_h \times d}$ are learned parameters, $d_h$ is the hidden dimension, and $C \in \mathbb{R}$ is a constant. This method has demonstrated its effectiveness in solving both polynomial and NP-hard problems, with a focus on generalization from small to large graphs.

\subsubsection{Monte Carlo Tree Search (MCTS)}
Monte Carlo Tree Search (MCTS) is a robust algorithm crucial in addressing complex decision spaces and combinatorial optimization problems. Operating through a systematic approach of four distinct stages, MCTS begins with \textit{Selection}, systematically navigating a search tree from the root node using a selection policy, typically the Upper Confidence Bound (UCB) for Trees, to balance exploration and exploitation. This stage is followed by \textit{Expansion}, where the algorithm extends the tree by creating child nodes to represent potential actions, effectively broadening the decision space. The process further proceeds to the \textit{Simulation (Rollout)} phase, where random actions are assessed from the selected node until reaching terminal states or a predefined depth. Finally, in the \textit{Backpropagation} phase, MCTS aggregates outcomes, updating statistical information such as visit counts and reward estimates for nodes along the path traversed during selection. This structured approach allows MCTS to systematically explore and update the search tree, avoiding local minima and effectively navigating complex decision spaces.

\begin{figure}
    \centering
    \includegraphics[width=0.7\linewidth]{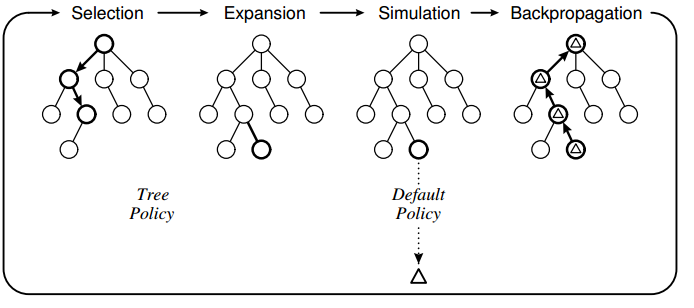}
    \caption{Illustration explaining Monte Carlo Tree Search}
    \label{fig:mcts}
\end{figure}

\section{Our Proposals}

This section delineate all the proposals and formulations presented in our article. Section \ref{section:pi_gnn} commences with an exposition of the early stopping strategy employed by PI-GNN as discovered in their work. Subsequently, we pinpoint a pivotal concern associated with this strategy, and thereafter, we introduce our modified strategy designed to improve the performance. Expanding on the concepts from the work of Drori et al. (2020) (see Section 2), which we refer to as GRL for brevity, Section \ref{section:grl} introduces our second contribution; a generic reinforcement learning (RL)-based framework that employs a QUBO-based Hamiltonian as the reward function, that we call $GRL_{QUBO}$. Here, we detail the modifications applied to the GRL framework, the elements retained, and our rationale for employing a QUBO-based reward. Finally, in Section \ref{section:mcts_gnn}, we present our thrid contribution, a Monty Carlo Tree Search-based solution, \textit{MCTS-GNN}, in addressing the CO problem based on manual perturbation of node labels to guide the search tree.

\subsection{PI-GNN with Fuzzy Strategy} \label{section:pi_gnn}
\textcolor{black}{PI-GNN uses Definition \ref{def:strict_stop} to perform early stopping during training. This intuitive support behind this strategy states that if we do not observe improvement or observe very insignificant improvement in the loss objective during training for a consecutive empirically set number of epochs, the training can be stopped. In general cases, this observation provides good output. But, as per our experimental observation, this strategy bears a crucial concern, especially in the graphs with higher densities.}

\begin{definition}[Strict Stopping] \label{def:strict_stop}
    During training, for the objective function $F_{obj}$, if no successive improvement is observed for consecutive $p$ epochs or the successive variation is lesser than $\tau$ occurs for consecutive $p$ epochs, the training can be stopped.
\end{definition}


During PI-GNN training, the objective or the loss function often starts with high positive value and then with gradual training moves into larger negative values. This phenomenon is common in graphs with high density. During the transition from positive to negative loss, there is often a period when the loss variation or improvement is minimal (e.g., less than $10^{-7}$, $10^{-8}$, etc.). As a result, the $\tau$ value or the early stopping criterion outlined in Definition \ref{def:strict_stop} may prompt an early halt to training, which can significantly diminish performance quality. Therefore, we propose the use of a fuzzy-stopping strategy as described in Definition \ref{def:fuzzy_stop}, which provides a more lenient stopping criterion. The rationale for the gradual loss variation is that nodes with a large number of connected edges tend to have feature vectors that are closely clustered in the vector space, leading to a uniform distribution of node probabilities. With additional iterations, these feature vectors are refined, resulting in more distinct patterns. It is also important to note that pinpointing an improved tolerance value during this gradual transition is challenging, rendering the early stopping approach in Definition \ref{def:strict_stop} less effective.

\begin{definition}[Fuzzy Stopping] \label{def:fuzzy_stop}
    Let us assume that, $\text{obj}^{*}$ denotes the current best value observed for the objective function $F_{\text{obj}}$ during any phase of the training iterations. If no improvement in the value of $F_{\text{obj}}$ occurs for successive $p$ epochs compared to $\text{obj}^{*}$, the training can be stopped.
\end{definition}
\hfill \newline

\textcolor{black}{It is readily apparent that Definition \ref{def:fuzzy_stop} provides a fuzzier criterion compared to Definition \ref{def:strict_stop}. This is because the latter one allows for both faster and slower transitions for the loss objective during training by removing the strict dependency over $\tau$.}

\subsection{A QUBO-based Generic Reinforcement Learning framework\:($GRL_{QUBO}$)} \label{section:grl}
We first discuss the modified portion of $GRL_{QUBO}$ that varies from $GRL$. Then, we point to the adopted portion brought into our architecture. The main purpose of experimenting with QUBO-based Hamiltonian as reward objective is generalization. A wide group of optimization problems has already been modeled in QUBO. Our research objective is to observe if we can use this QUBO-based model as a generic reward function for RL-based formulations in a permutation-invariant manner. This contribution works as a bridge between the QUBO-based CO solutions with RL-based approaches.


\begin{enumerate}
    \item \textbf{Generic QUBO-formulated Hamiltonian Reward Function:} A subset of terms from $X^{T}QX=\sum_{i \leq j}x_{i} Q_{ij}x_{j}$ is considered the observed reward $r^{t}$ at time $t$ during training for a particular epoch $e$. When a node $v_{i}$ is greedily selected and labeled, we check which terms $x_{i}Q_{ij}x_{j}$ where $i \leq j$ and $x_{j}Q_{ji}x_{i}$ where $j < i$ can be calculated and sum them. This is considered as the reward, $r^{t} = \sum_{i \leq j} x_{i}Q_{ij}x_{j} + \sum_{j < i}x_{j}Q_{ji}x_{i}$.
    \item \textbf{Attention-based decoding strategy:} In \textcolor{black}{Equation }\ref{eqn:decoder_attn_modified}, we provide the mathematical formulation to calculate the attention or weight for node $v_{i}$ as $\alpha_{i}$. Here, $\phi_{1} \in \mathbb{R}^{d_{h} \times d}, \phi_{2}\in \mathbb{R}^{d_{h} \times d}$ and $\phi_{3}\in \mathbb{R}^{d_{h} \times n}$ are weights or architecture parameters. $h_{i}$ denotes the node feature vector (row vector) for the node $v_{i}$ reported from the GAT layer. $C$, $d$ and $d_{h}$ all are hyperparameters. $n$ denotes the number of nodes in the input graph. Through applying a sigmoid non-linear activation over $\alpha_{i}$, node probability distribution $P_{i}(\theta)$ is generated. Over $P_{i}(\theta)$, a probability threshold $\beta$ ($e.g., \beta=0.5$) is applied to fix the node labels, e.g., $P_{i}(\theta) \geq \beta \text{ leads to } x_{i}=1$. Here $\theta$ denotes the complete set of trainable architecture parameters and $T$ is used to denote the transpose of a vector (row to column or column to row).

    \begin{equation}
    \label{eqn:decoder_attn_modified}
    \alpha_{i} =  C \text{ tanh}(\frac{{ h_{i}\phi_{1}^{T} .(X_{v}\phi_{3}^{T} + \sum_{j \in N(i)} h_{j}\phi_{2}^{T}) }}{\sqrt{d_{h}}})
\end{equation}

    \textcolor{black}{We incorporate more context in Equation \ref{eqn:decoder_attn_modified} compared to Equation \ref{eqn:grl_decode} to update a node $v_{i}$'s attention weight $\alpha_{i}$. \textcolor{black}{Equation \ref{eqn:grl_decode} is the decoding strategy of $GRL$ as discussed in Section \ref{sec:rlco}. }In Equation \ref{eqn:grl_decode}, based on the selected node $v_{i}$ we update all of its adjacent node $v_{j}$'s attention values ($\alpha_{ij}$) over one to one dependencies. But, in Equation \ref{eqn:decoder_attn_modified}, to update a node $v_{i}$'s attention value, we consider all of its adjacent nodes, $v_{j}$'s in the same time along with the status of already labeled nodes. To be precise, in Equation \ref{eqn:decoder_attn_modified}, we consider three aspects - node $v_{i}$'s feature vector ($h_{i}$), all of its adjacent neighbor's ($N(i)$) feature vectors (($h_{j} \text{ where } j \in N(i)$)) and the current condition of node labeling $X_{v}$. If a node $v_{j}$ has already been selected and labeled then the $j^{th}$ entry will be $1$ otherwise it will remain as $0$. So $X_{v}$ is a binary vector, $\{0, 1\}^{n}$. Here $n$ denotes the number of nodes in the graph.}

\end{enumerate}

Now, we point out the common strategies that we adopt from $GRL$ through the following points,

\begin{enumerate}
    \item \textbf{GAT as an encoder architecture:} $K$ stacked layers of GAT are used as the encoder architecture to generate the node feature vectors $h$. The last layer consists of a sigmoid non-linear activation layer to generate the node probabilities. This probability helps to select the first node to initiate the decoding with the attention mechanism.
    \item \textbf{Loss objective and Training:} The loss objective that we use is stated in \textcolor{black}{Equation }\ref{eqn:decoder_attn_modified}. Here $P(v^{t})$ denotes the probability of the greedily selected node $v$ at the $t^{th}$ iteration. $r^{t}$ has already been defined prior. $b$ denotes the reward observed from the baseline architecture at the $t^{th}$ iteration. \textcolor{black}{As per the source paper of \textcolor{black}{GRL}, baseline architecture can be considered a result produced from a problem specific heuristic solution or a sub-optimal solution. The source paper tries to incorporate this heuristically produced value at each time stamp of the reward function or loss function. In our setup. for the simplicity and generalization point of view, we omit this term and depend only on the QUBO-based Hamiltonian. As per our empirical results, we still observed quite comparative performance which will be demonstrated in Section \ref{section:evaluation}.} After accumulating the complete set of rewards (termination of an epoch), we apply gradient descent over the model parameters and backpropagate.

    \begin{equation}
        L(\theta) = \sum_{t=1}^{n} (r^{t}-b) \times P(a^{t})
         =  \sum_{t=1}^{n} r^{t} \times P(a^{t}) [\text{when $b=0$}]
    \end{equation}
\end{enumerate}

\subsection{Monty Carlo Tree Search with GNN through manual perturbation, MCTS-GNN} \label{section:mcts_gnn}
In this section, we present a formulation where we integrate the Monty Carlo Tree Search in assistance with GNN in an RL-based setup. The main idea is that each node of the search tree consists of a partial solution (or node labels), and based on that a single GNN is trained to approximate the remaining nodes' labels. The goal is, using this manual perturbation of node labels, a guided search is conducted to maximize the amount of rewards. Now, we discuss the strategies based on RL terminologies (state, action, reward) and Monty Carlo tree search terminologies (selection, rollout, exploration, and backpropagation) in a brief manner.

\begin{enumerate}
    \item \textit{State} $S$: \space Each node or a state $S$ of the MCTS tree, provides a partial solution or a subset of possible node labeling for the concerned CO problem. Similar to the previous section, for the processing we maintain a binary vector $X_{v}$ where the $i^{th}$ entry is set to $1$ if $i^{th}$ node has already been labeled.
    \item \textit{Action}, $a$ and \textit{Transition function}, $\pi(S, a)$: \space An action $a$ means choosing a label (either $0$ or $1$) for the input graph node variable $x$. From each state $S$ multiple actions can be created by fixing the node labels of the unselected nodes from the point of view of $S$. Each action $a$ from the state $S$ also bears a transition probability $\pi(S, a)$ denoting the likelihood of taking action $a$ from $S$. $\pi(S,a)$ is approximated using a GNN.
    \item \textit{Reward}, $r$: \space To calculate the reward for a state $S$, GNN is used. Using GNN, node probability distributions $P(\theta)$ are generated. Based on $S$, a subset of nodes have already been fixed, for the remaining unselected nodes' labels, $P(\theta)$ is used over a probability threshold $\beta$, e.g., for a node $v_{i}$, if $P_{i}(\theta) \geq \beta$, then $x_{i}=1$ else $x_{i}=0$. After approximating the labels of all the nodes, QUBO-formulated Hamiltonian is used to calculate the reward, $r = \sum_{i \leq j} x_{i}Q_{ij}x_{j}$.
\end{enumerate}

As it \textcolor{black}{is already obvious}, GNN plays a very important part of this design, now we state the mathematical formulation of GNN's forward pass along with the loss function that is used to update the parameters $\theta$.

  \begin{equation}
     \label{eqn:x_em}
         X_{em} = E(G)
    \end{equation}
    \begin{equation}
         \label{eqn:mu}
        h^{\prime} = GNN(G, X_{em})
    \end{equation}
    \begin{equation} \label{eqn:mu_final}
        h = f_{1}(X_{v}\theta_{1}^{T} + h^{\prime}\theta_{2})
    \end{equation}
    \begin{equation}
        \label{eqn:predict_mcts_gnn}
         P(\theta) = f_{2}(h\theta_{3}^{T})
    \end{equation}

The complete set of the mathematical formulation is presented from \textcolor{black}{ Equation~\ref{eqn:x_em}} to \textcolor{black}{Equation~\ref{eqn:predict_mcts_gnn}}. First,
a set of node embedding vectors $X_{em}$ is generated (\textcolor{black}{Equation }\ref{eqn:x_em}). Then, $X_{em}$ and input graph G is passed to GNN to generate
a set of node feature vectors $h^{\prime}$ (\textcolor{black}{Equation} \ref{eqn:mu}). After that, more contextual information ($X_{v}$) is added with feature vector ($h^{\prime}$) to calculate the complete node feature vectors $h$ (\textcolor{black}{Equation } \ref{eqn:mu_final}). \textcolor{black}{Equation \ref{eqn:predict_mcts_gnn} represents the final equation to generate the node probability distribution $P(\theta)$}. Here the set of architecture parameters $\theta = \{\theta_{GNN} \in \mathbb{R}^{d_2 \times d_1}, \theta_{1} \in \mathbb{R}^{d_3 \times 1}, \theta_{2} \in \mathbb{R}^{d_3 \times d_2}, \theta_{3} \in \mathbb{R}^{ 1 \times d_3} \}$. We use $\theta_{GNN}$ to denote all the weight parameters associated with the layers of GNN. $f_{1}$ is a ReLU activation function and $f_{2}$ is a sigmoid activation function to generate the probabilities. Similar to before
$(.)^{T}$ denotes the transpose from row vector to column vector or vice versa. $P(\theta)$ is also used to approximate $\pi(S, a)$. For a particular variable $x_{v}$ or input graph
node $v$, $P_{v}(\theta)$ will indiciate the likelihood of labeling $v$ to $1$ ($x_{v}=1, \pi(S, v=1)=P_{v}(\theta)$). Similarly, to label $x_{v}$ as $0$, we set $\pi(S,v=0)=1-P_{v}(\theta)$. From each state $S$, we create child nodes for the unselected variables for both of the labels.

To train the GNN architecture\textcolor{black}{,} we use the formulation stated in \textcolor{black}{Equation} \ref{eqn:loss_mcts_gnn} as the loss function. This function is quite similar to \textcolor{black}{Equation \ref{eqn:pi_gnn_loss} apart from the fact} that here we add manual perturbation by fixing the node labels to guide the searching. Here $X_{v,i}$ denotes the value of the $i^{th}$ entry in $X_{v}$. \textcolor{black}{Here, $X_{v,i}$ is set to $1$ when node $i$ has already been labeled. On the contrary, $X_{v,i}$ is set to $0$ when the node $v_{i}$ has not been labeled.} Our underlying intuition behind this formulation is presented in the form of the Proposition \ref{proposition:manual_perturb}.

 \begin{equation} \label{eqn:loss_mcts_gnn}
        L(\theta) =  \sum_{i \leq j, X_{V,i}=1, X_{v,j}=1} x_{i} Q_{ij}x_{j} + \sum_{i \leq j, X_{V,i}=1, X_{v,j}=0} x_{i} Q_{ij} p(\theta_{j}) + \sum_{i \leq j,X_{V,i}=0, X_{v,j}=0} p(\theta_{i}) Q_{ij} p(\theta_{j})
    \end{equation}
\textcolor{black}{
\begin{proposition}[Manual perturbation to avoid local minimas] \label{proposition:manual_perturb}
The relevant arguments related to this proposition are given by the following points,
\begin{itemize}
    \item A single GNN is trained through different sets of manually set node labels(partial solutions) which we regard as manual perturbation. These sets of possible partial solutions can be considered as noises or data points.
    \item The GNN during its training updates its weight parameters by tackling these sets of perturbations.
    \item This strategy mainly forces the GNN to adapt itself to different variations which as per our intuition relates to learning strategies to avoid or push itself from various local minimas observed during training. This results in more robust architecture and improved performance in terms of reducing constraint violations.
\end{itemize}
\end{proposition}}
\hfill \break

Now, we discuss the terminologies associated with MCTS through the following points,

\begin{enumerate}
	\item \textbf{Selection and Exploration:} In \textcolor{black}{Equation }\ref{eqn:selection}, we present the greedy metric, Upper Confidence Bound (UCB) to measure the average reward obtainable for a child state $C_{i}$ with respect to its parent state $S$ combining its transition likelihood ($\pi$) of being selected. Here $C_{i}.w$ and $C_{i}.v$ denote the total amount of reward accumulated in state $C_{i}$ and the number of times $C_{i}$ has been visited, respectively. $\alpha$ denotes a hyperparameter, $\pi(S, a)$ denotes the transition probability to state $C_{i}$ from $S$ reported by GNN. $S.v$ denotes the total number of times state $S$ was visited. $\text{log}$ denotes mathematical logarithmic operation. The node with the highest UCB value is selected to explore in its subtree. When a leaf node is reached in this manner, we start the rollout phase.

         \begin{equation} \label{eqn:selection}
            UCB(C_{i})= \frac{C_{i}.w}{C_{i}.v} + \alpha * \pi(S, a) * \sqrt{\frac{\text{log}(S.v)}{C_{i}.v}}
    \end{equation}

	\item \textbf{Rollout:} In the rollout phase, using GNN, we conduct the training for multiple epochs to approximate the node labels for the unlabeled nodes and calculate the rewards. We have already discussed how GNN is used to generate the node probability distributions in the previous paragraphs.
	\item \textbf{Backpropagation:} After the rollout phase we enter the backpropagation phase of MCTS. In this phase, we update the state variables, $v$ and $w$ for the path from the current leaf state to the root of the search tree. For all the nodes in the path we increment the visiting attribute by $1$ and add the reward to $w$ approximated by GNN.
\end{enumerate}

\section{Evaluation} \label{section:evaluation}


In this section, we empirically evaluate $GRL_{QUBO}$ and MCTS-GNN with PI-GNN in addressing the Max-Cut problem for the graphs of different densities. The graphs were randomly generated and all the graphs were undirected. For a given graph of $n$ nodes, there can be at most $\frac{n \times (n-1)}{2}$ edges where any two nodes do not have multi-edges among them. Our prepared graph dataset can be found in \footnote{\url{}}. A statistics over the graphs based on their adjacency information is shown in Table \ref{tab:graph_dist} denoting the maximum, minimum and average number of adjacent edges for each of the nodes for all the experimented graphs. We conduct the experiments based on three metrics - the number of satisfied constraints, \textcolor{black}{training stability}, and required time. All the experiments were conducted on a 64-bit machine having AMD Ryzen 9 5950x 16-Core Processor x 32, 128 GB RAM, and 24 GB NVIDIA GeForce RTX 3090 GPU. All the implementations were done in Python language based on the blocks provided by deep-learning libraries Pytorch \citep{paszke2019pytorch}, Pytorch Geometric \citep{fey2019fast} and LabML \citep{labml}.

\begin{table}[]
    \centering
    \begin{tabular}{c|c|c|c|c}
        node & edges & Maximum number of & Minimum Number of & Average Number of\\
        & & Adjacent edges & Adjacent edges  & Adjacent edges \\
        \hline
        50 & 89 & 11 & 1 & 3.56\\
        50 & 139 & 10 & 2 & 5.56\\
        50 & 499 & 27 & 13 & 19.96\\
        \hline
        100 & 199 & 11 & 1 & 3.98\\
        100 & 799 & 24 & 8 & 15.98\\
        \hline
        300 & 399 & 7 & 1 & 2.66\\
        300 & 899 & 13 & 1 & 5.99\\
        300 & 1299 & 18 & 1 & 8.66\\
        \hline
        500 & 799 & 10 & 1 & 3.2\\
        500 & 1499 & 13 & 1 & 6.0\\
        500 & 5499 & 45 & 10 & 22.0\\
        \hline
        700 & 1199 & 9 & 1 & 3.43\\
        700 & 1699 & 13 & 1 & 4.85\\
        700 & 4699 & 24 & 4 & 13.43\\
        \hline
        1000 & 1299 & 8 & 1 & 2.6\\
        1000 & 3299 & 16 & 1 & 6.6\\
        1000 & 5299 & 21 & 1 & 10.6\\
        \hline
        3000 & 3499 & 8 & 1 & 2.33\\
        3000 & 4499 & 9 & 1 & 3.0\\
        3000 & 6999 & 13 & 1 & 4.67\\
        \hline
    \end{tabular}
    \caption{Graph Distribution Statistics}
    \label{tab:graph_dist}
\end{table}

\subsection{Architecture Description, Hyperparameter Setup}
In this section, we present the description of the architectures along with the value of the hyperparameters that are used during the training. The description is presented in the following manner,

\begin{enumerate}
    \item \textbf{Architecture Description:} To present the discussion we mostly use the variables stated in the respective sections.

    Our GNN architecture follows \textcolor{black}{the similar definition provided in PI-GNN}. It consists of two layers of Graph Convolutional Architecture (GCN) and a sigmoid layer to generate the node probabilities. The first GCN layer's node feature vectors are propagated through a non-linear ELU activation and a dropout layer before passing to the second layer of GCN. Let us assume the node feature vector size of the $1^{st}$ and $2^{nd}$ layers of GCN are $d_{1}$ and $d_{2}$. As mentioned in the base article, we follow a similar setup. If the number of nodes $n \geq 10^{5}$, then $d_{1}=\sqrt[3]{n}$, else $d_{1}=\sqrt{n}$. Also, $d_{2} = \frac{d_1}{2}$.

    To implement $GRL_{QUBO}$\textcolor{black}{,} we took inspiration from the study presented \textcolor{black}{in the base source paper \textcolor{black}{of GRL}}. Here, we had three layers of GAT as the encoding architecture and a sigmoid layer to generate initial node probabilities. The attention-based decoding formulation has been presented in Section \ref{section:grl}. To set the dimension of the node feature vectors, we follow a similar strategy stated in the previous paragraph based on the number of nodes of the input graph, $n$. So, if $n \geq 10^{5}$ then, $d_{1}=\sqrt[3]{n}$ else, $d_{1}=\sqrt{n}$ and followup $d_2 = \lceil \frac{d_1}{2} \rceil$. We have used a single attention-head mechanism.

    In MCTS-GNN, we train a single GNN in different rollout phases by applying different manual labeling perturbations. The setup is completely similar to the discussion already presented above. So, if $n \geq 10^{5}$, then  $d_{1}=\sqrt[3]{n}$ else, $d_{1}=\sqrt{n}$. $d_2 = \lceil \frac{d_1}{2} \rceil$ and $d_{3}=1$.
     \item \textbf{Training Description:} To train the architectures, we use Adam optimizer for each of the architectures. We use patience value to implement fuzzy early stopping for all of them. As RL-based setups inherently exhibit abrupt behavior (frequent ups and downs in terms of reward variation) we keep a comparatively higher patience value for $GRL_{QUBO}$ and MCTS-GNN compared to PI-GNN. As per our setups, we eventually observed almost linear variation in terms of objective functions' values at some phase of the training epochs for all the experimented architectures for different graph inputs. This supports the validity of the chosen hyperparameters for early stopping. \textcolor{black}{There lies another reason behind this strategy. In PI-GNN, we observe the variation in the loss values which is fractional whereas in the other experimented architectures we track the reward values which is integer. Inherently, it can be understood that the integer rewards should fluctuate more compared to the fractional loss values. This demands a higher early stopping threshold for the RL-based setups.}

     For PI-GNN, we simply choose learning rate $lr=10^{-4}$ with a patience value of $\tau=100$ denoting if there is no loss objective improvement compared to the current best value observed for a consecutive $100$ epochs, we stop the training. As stated previously, we apply fuzzy early stopping \textcolor{black}{in our experimented setup}. For $GRL_{QUBO}$, we choose a learning rate of $0.001$ ($lr$) for both encoder-decoder and a patience value of $700$ ($\tau$). We also experimented with a lesser learning rate ($0.0001$) for $GRL_{QUBO}$ \textcolor{black}{but could not observe further improvement} in terms of satisfying the number of constraints. For, MCTS-GNN, we also chose the learning rate $lr$ to $0.001$ for the GNN architecture with the patience value of $100$. \textcolor{black}{Apart from the prior hyperparameter to control GNN training, we included another early stopping criterion $\tau^{\prime}$ that tracked the improvement in the reward objective. We set $\tau^{\prime}$ to $700$  denoting that if there is no reward improvement compared to the best reward observed till the current iteration, the MCTS-GNN algorithm terminates.}
\end{enumerate}

\subsection{Comparison in terms of number of satisfied constraints}
In this section, we present the number of satisfied constraints observed in each architecture, PI-GNN, $GRL_{QUBO}$ and MCTS-GNN, for the graphs of different intensities. The complete result is shown in Table \ref{tab:reward}. In this study, this metric is denoted as the reward in terms of RL-based formulation. For $GRL_{QUBO}$ and MCTS-GNN we present the best reward observed throughout the complete training. For PI-GNN, we record the least loss and the corresponding probability distribution to approximate the node labels based on the threshold \textcolor{black}{$\beta$( e.g. if $\beta=0.5$, then the node's probability exceeding $0.5$ gets labeled as $1$)}. These approximated node labels are used to calculate the QUBO-formulated Hamiltonian function. In the last two columns of Table \ref{tab:reward}, we present how $GRL_{QUBO}$ and MCTS-GNN's performance has improved compared to PI-GNN in percentage. A positive value means, the result has improved and the negative means the opposite.

\begin{table}[!ht]
    \centering
    \begin{tabular}{c|c|c|c|c|c|c}
    \hline
    \# Nodes & \# Edges & PI-GNN & $GRL_{QUBO}$ & MCTS-GNN & $GRL_{QUBO}$ & MCTS-GNN\\
    &  & & & &  vs PI-GNN(\%)  &  vs PI-GNN (\%) \\
    \hline
    50 & 89 & 72 & 75 & 76* & 4.2 & 5.56\\
    50 & 139 & 95 & 105 & 107 * & 10.53 & 12.63 \\
   50 & 499 & 276 & 301 & 314* & 10.14 & 13.78\\
   \hline
100 & 199 & 147 & 155 & 167* & 5.44 & 13.61\\
100 & 799 & 373 & 534 & 537* & 43.2 & 44\\
\hline
300 & 399 & 342 & 360 & 365* & 5.26 & 6.73\\
300 & 899 & 613 & 690 & 699* & 12.56 & 14.03\\
300 & 1299 & 747 & 932 & 947* & 24.77 & 26.77\\
\hline
500 & 799 &  622 & 672 & 694* & 8.04 & 11.58\\
500 & 1499 & 1007 & 1143 & 1158* & 13.51 & 15.0\\
500 & 5499 &  2689 & 3414 & 3535* & 26.96 & 31.46\\
\hline
700 & 1199 &   938 & 979 & 1023* & 4.37 & 9.06\\
700 & 1699 &  1288 & 1308 & 1360* & 1.55 & 5.59\\
700 & 4699 &  2422 & 2992 & 3202* & 23.53 & 32.2\\
\hline
1000 & 1299 & 1104 & 1112 & 1141* &  0.73 & 3.35\\
1000 & 3299 & 2204 & 2449 & 2525* & 6.9 & 14.56\\
1000 & 5299 & 3098 & 3716 & 3750* & 19.95 & 21.05\\
\hline
3000 & 3499 & 2956 & 3218* & 2996 & 8.86 & 1.35\\
3000 & 4499 & 3627 & 3907* & 3885 & 7.72 & 7.11\\
3000 & 6999 & 4841 & 5550 & 5622* & 14.65 & 16.13\\
    \hline

    \end{tabular}
    \caption{Number of Satisfied Constraints (Reward) for the graphs of different intensities for the Max-Cut problem. (*) denotes the best value observed.}
    \label{tab:reward}
\end{table}

From the presented result in Table \ref{tab:reward}, we can see that, the performance has comparatively improved than the base PI-GNN. In simple terms, upon applying RL-based formulation we improved the quality of the resultant output. If we investigate in a more detailed manner we can see that, for a particular graph having $n$ nodes, the performance generally improves more with it being denser or increased number of edges.

As per our analysis, PI-GNN is quite scalable and a generic solution to address a wide range of CO problems. However, \textcolor{black}{it lags behind RL-based methods in meeting constraints, especially as the graphs get denser.} In $GRL_{QUBO}$, when the attention-based decoding scheme selects a particular node and sets its label, only its adjacent unlabeled neighbor nodes' attention weights ($\alpha$) are updated. Then using a max heap-based strategy next likely node is selected and labeled. As, in each selection, the binary vector is updated, the performance might be improved if all the nodes' attention weights are updated. However, it will add a significant runtime cost due to linear updates in each selection, so a runtime-accuracy tradeoff exists there. MCTS-GNN, intuitively applies noise during training to enforce the training or the architecture to skip various local minimas to reach comparatively better outcomes.

\subsection{Training Stability}
\begin{figure}
    \centering
    \begin{subfigure}[b]{0.32\textwidth}
        \includegraphics[width=\textwidth]{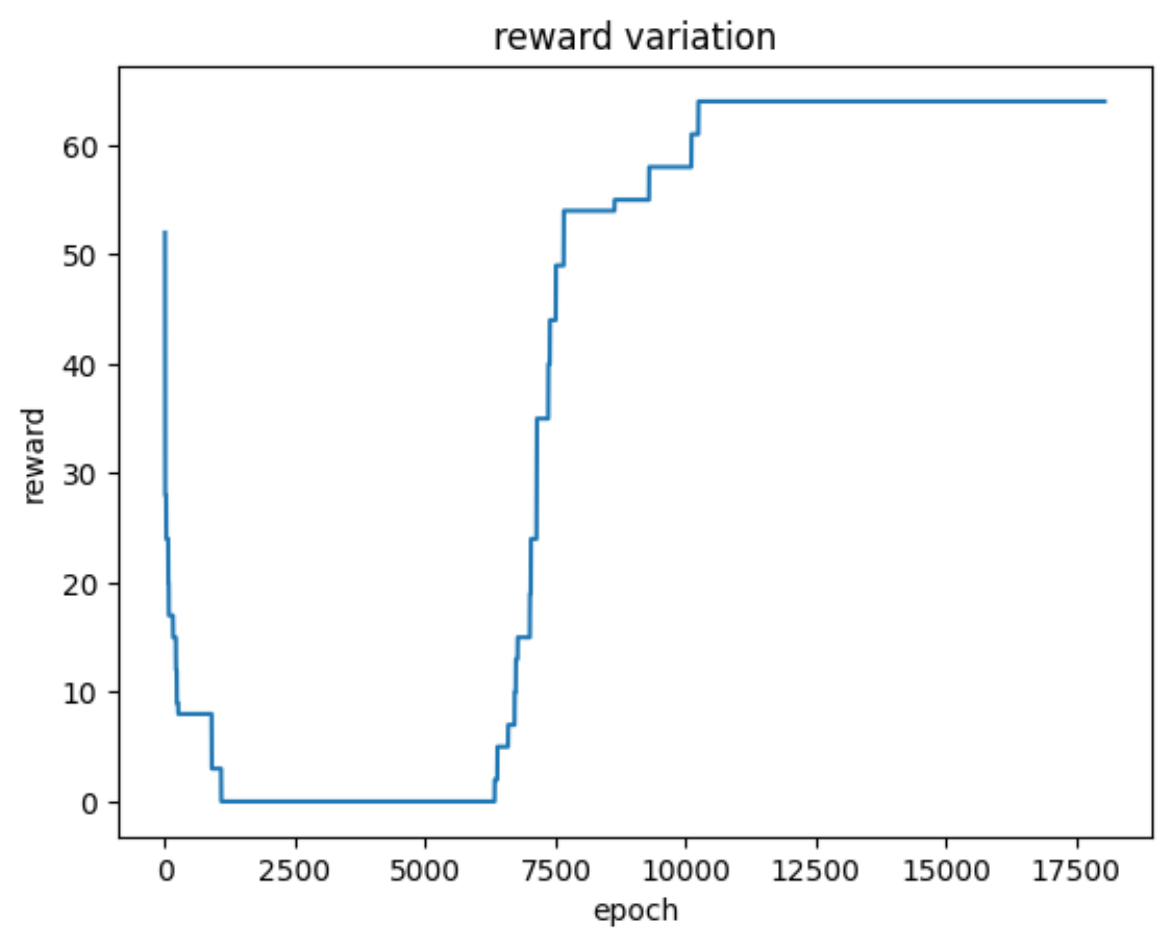}
        \caption{PI-GNN}
    \end{subfigure}
    \begin{subfigure}[b]{0.32\textwidth}
        \includegraphics[width=\textwidth]{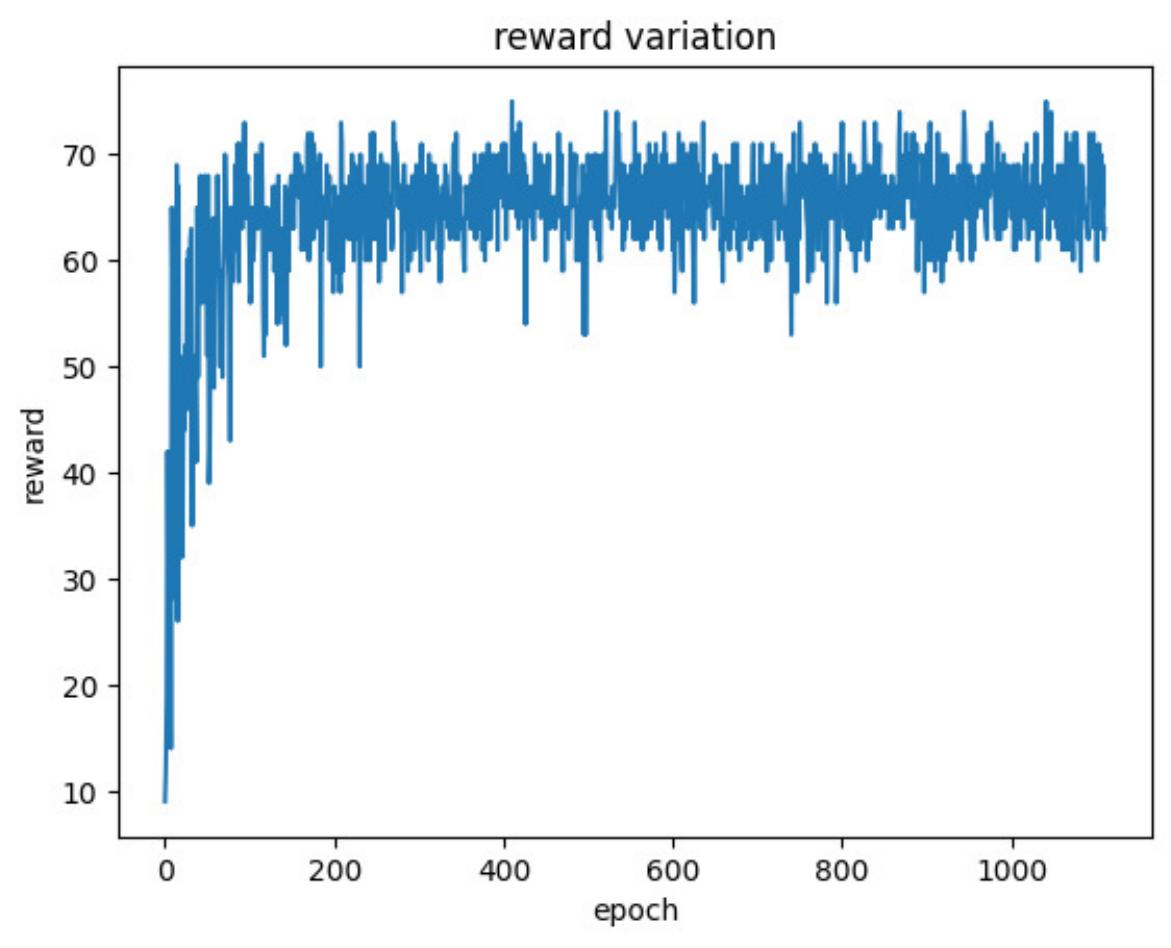}
        \caption{$GRL_{QUBO}$}
    \end{subfigure}
    \begin{subfigure}[b]{0.32\textwidth}
        \includegraphics[width=\textwidth]{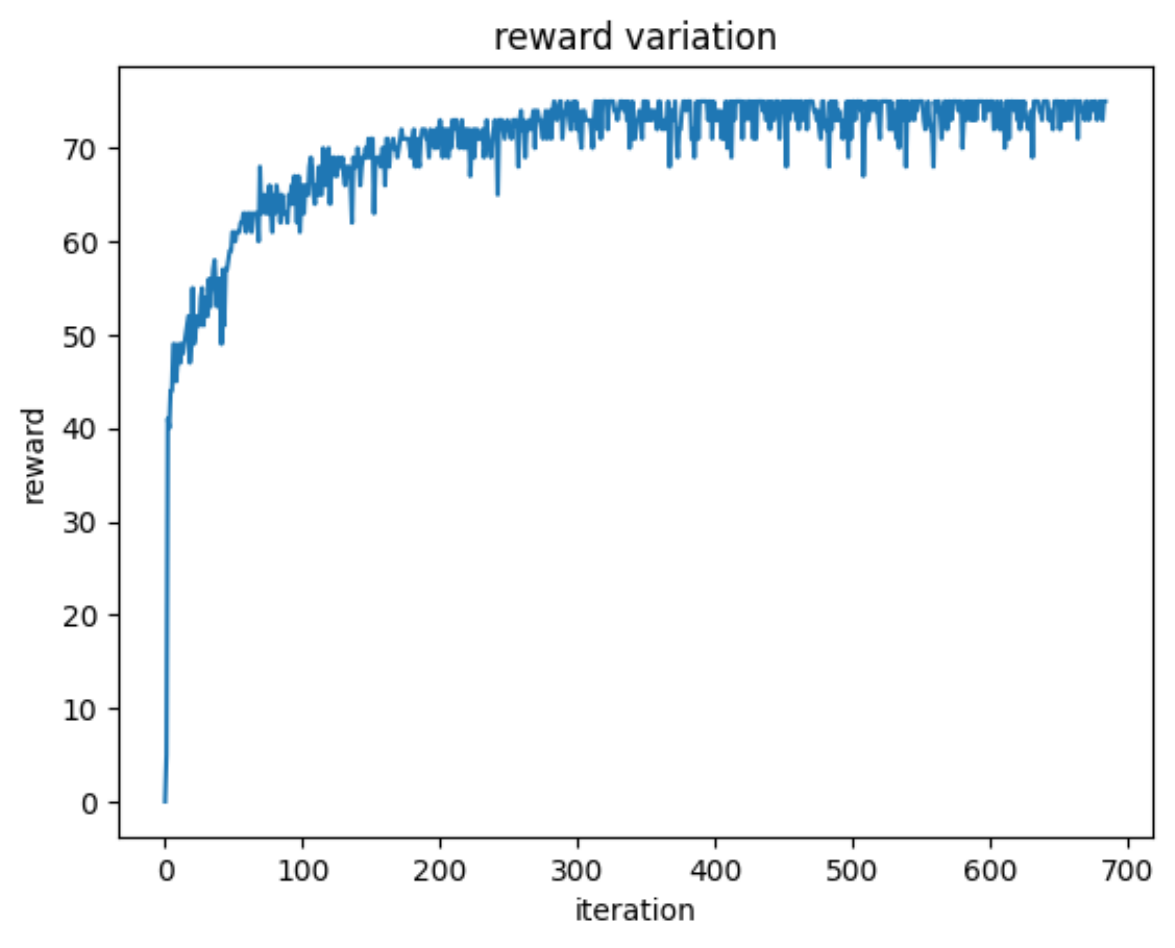}
        \caption{MCTS-GNN}
    \end{subfigure}
    \caption{Reward variation curve for PI-GNN, $GRL_{QUBO}$ and MCTS-GNN for the graph (50, 89)}
    \label{fig:50_89}
\end{figure}

\begin{figure}
    \centering
    \begin{subfigure}[b]{0.32\textwidth}
        \includegraphics[width=\textwidth]{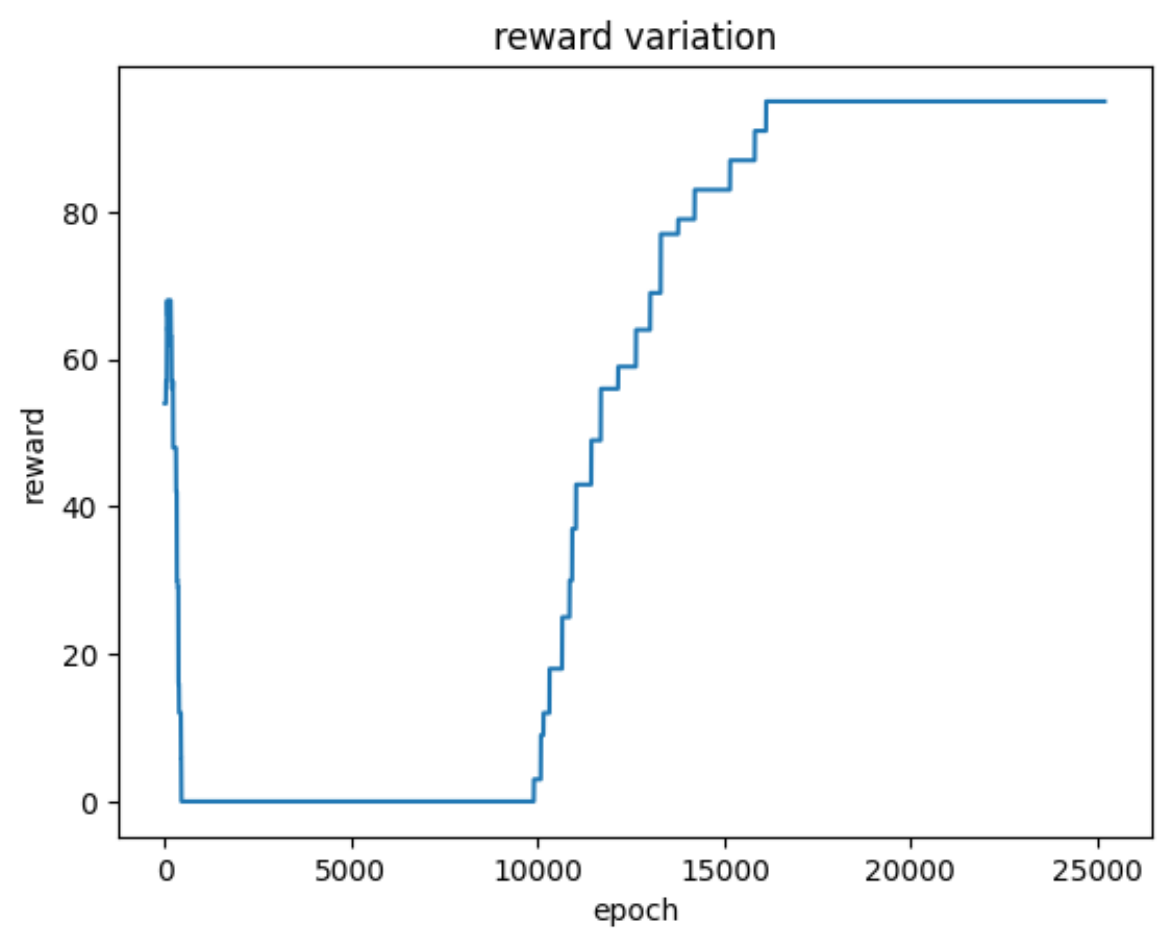}
        \caption{PI-GNN}
    \end{subfigure}
    \begin{subfigure}[b]{0.32\textwidth}
        \includegraphics[width=\textwidth]{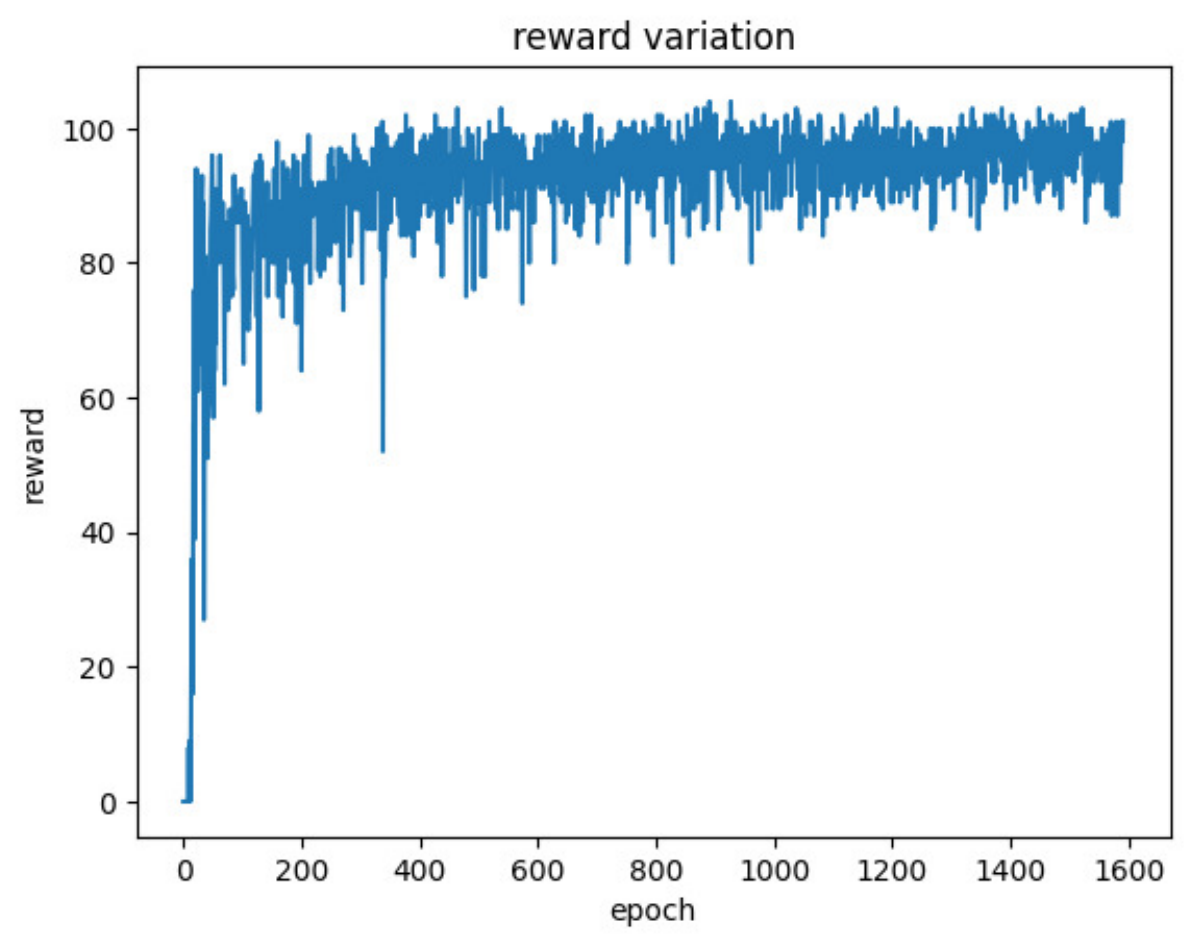}
        \caption{$GRL_{QUBO}$}
    \end{subfigure}
    \begin{subfigure}[b]{0.32\textwidth}
        \includegraphics[width=\textwidth]{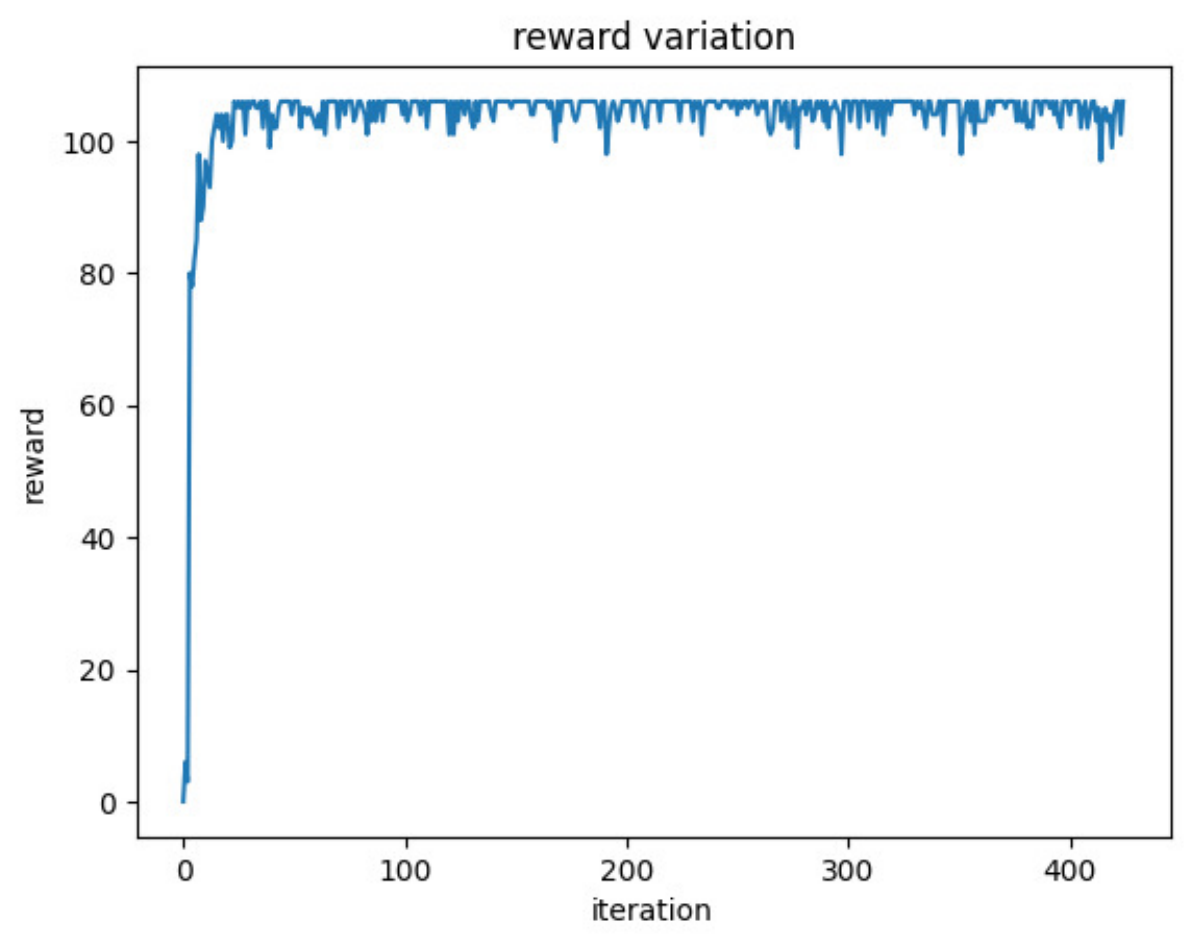}
        \caption{MCTS-GNN}
    \end{subfigure}
    \caption{Reward variation curve for PI-GNN, $GRL_{QUBO}$ and MCTS-GNN for the graph (50, 139)}
    \label{fig:50_139}
\end{figure}

\begin{figure}
    \centering
    \begin{subfigure}[b]{0.32\textwidth}
        \includegraphics[width=\textwidth]{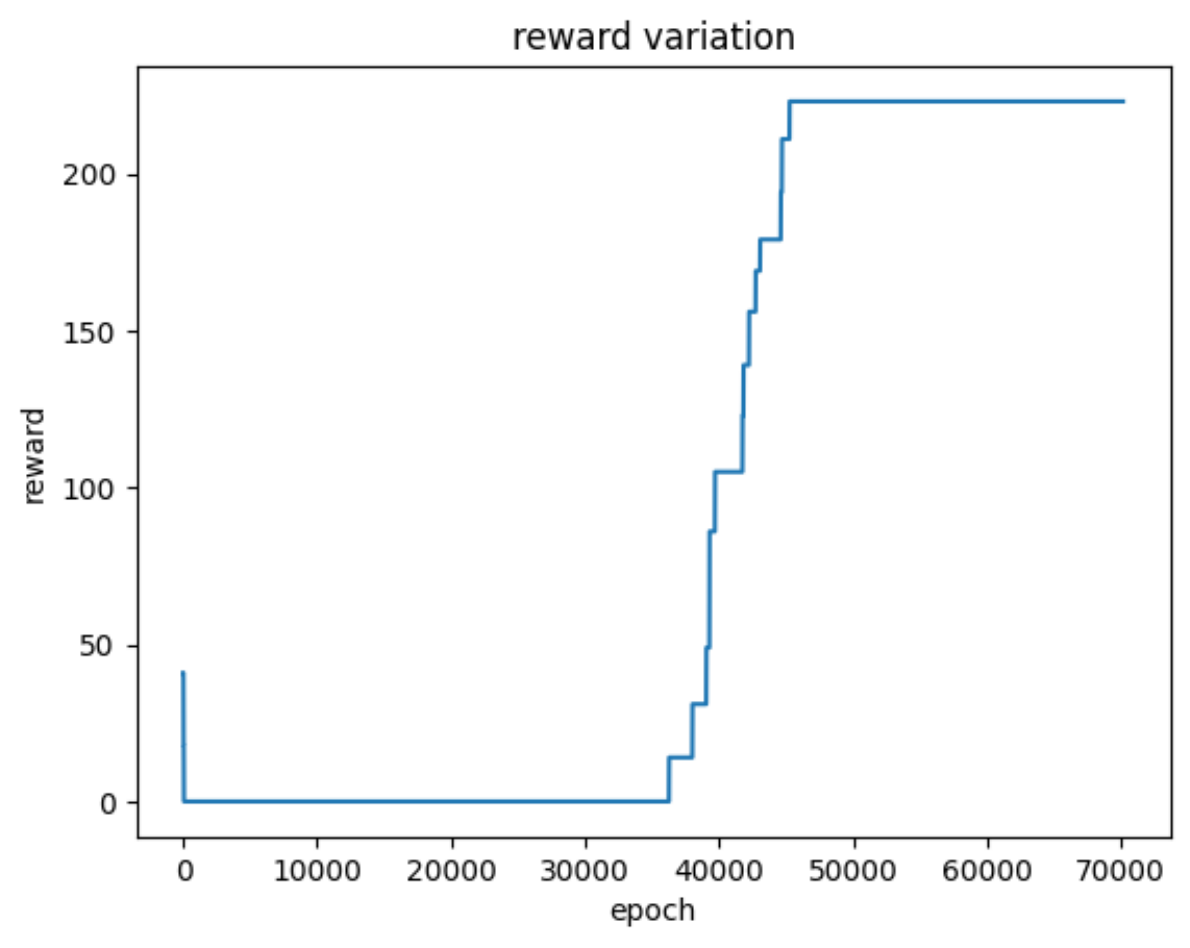}
        \caption{PI-GNN}
    \end{subfigure}
    \begin{subfigure}[b]{0.32\textwidth}
        \includegraphics[width=\textwidth]{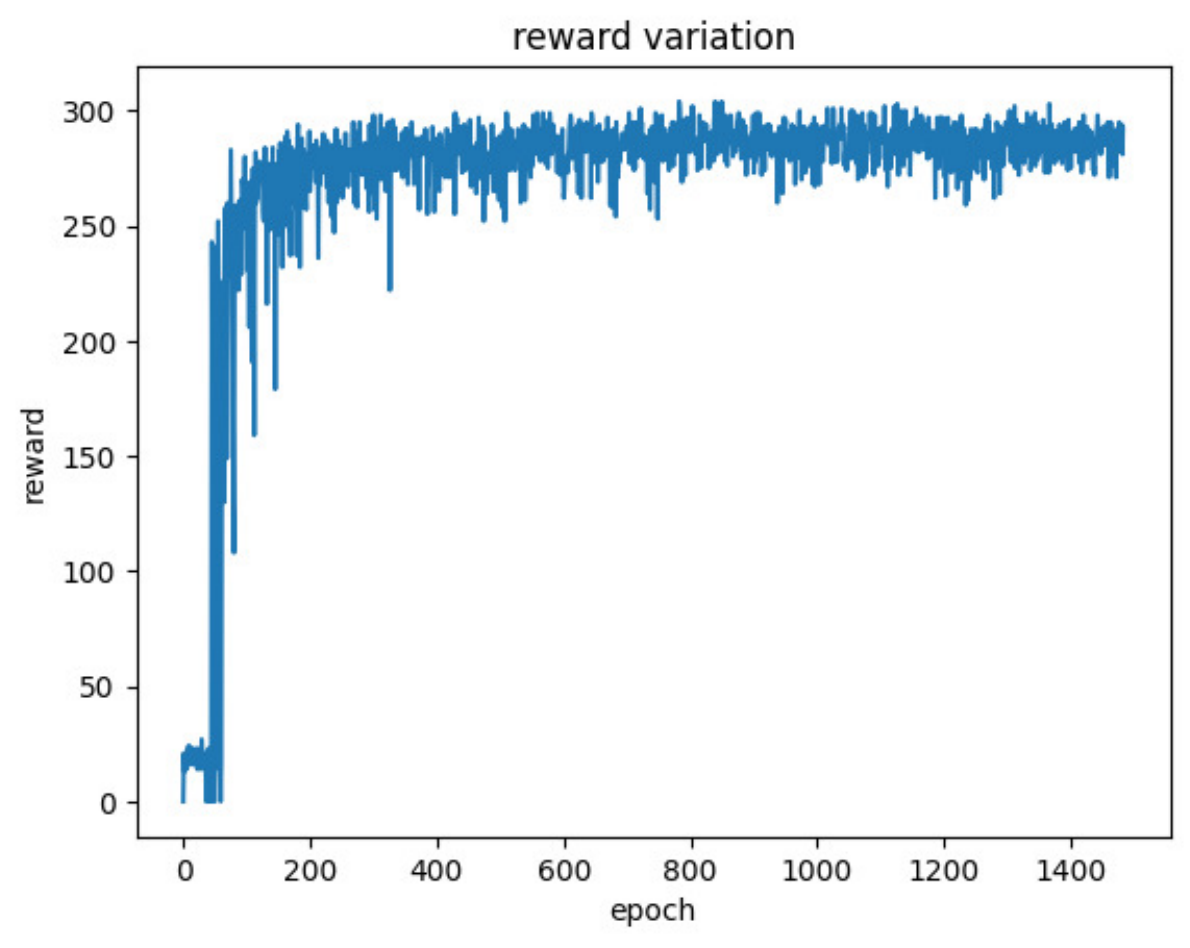}
        \caption{$GRL_{QUBO}$}
    \end{subfigure}
    \begin{subfigure}[b]{0.32\textwidth}
        \includegraphics[width=\textwidth]{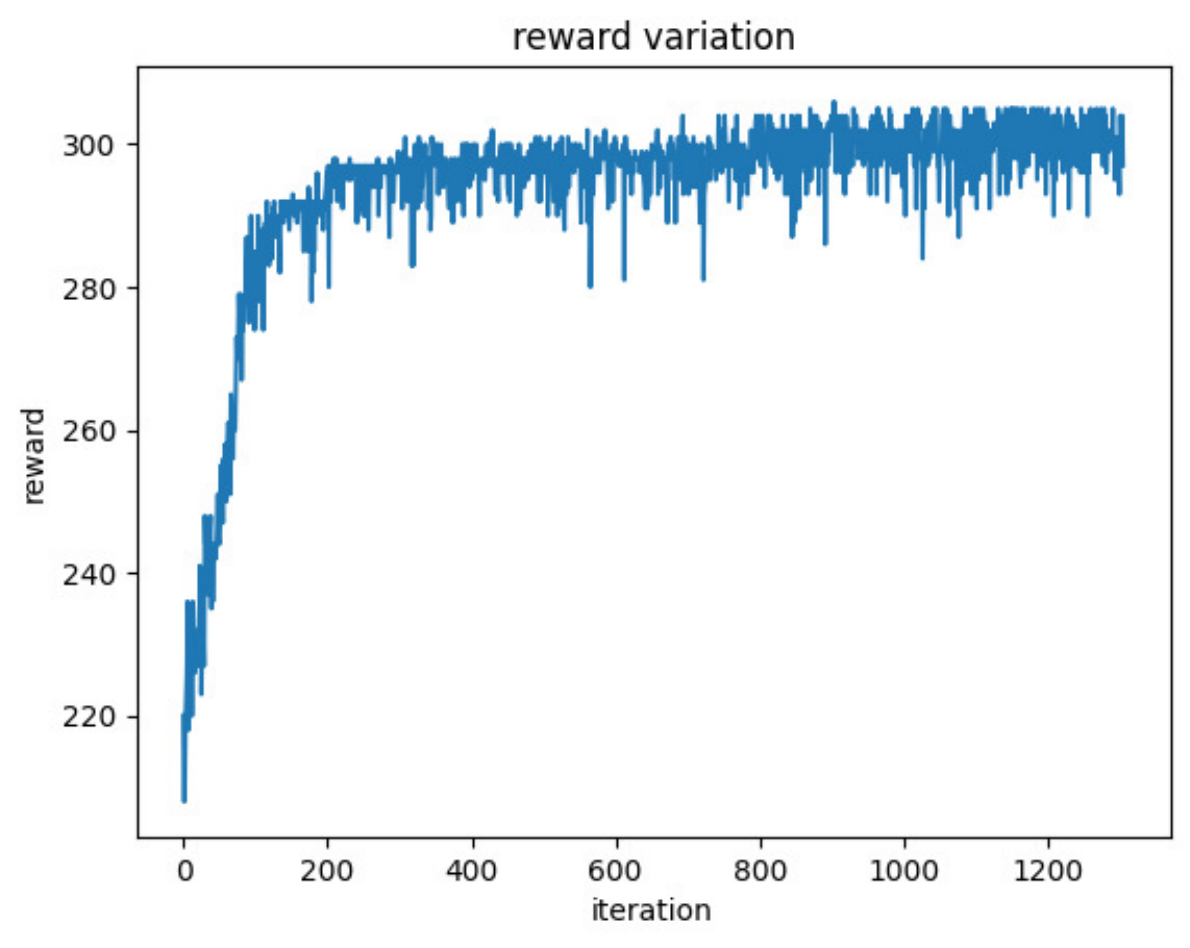}
        \caption{MCTS-GNN}
    \end{subfigure}
    \caption{Reward variation curve for PI-GNN, $GRL_{QUBO}$ and MCTS-GNN for the graph (50, 499)}
    \label{fig:50_499}
\end{figure}

\begin{figure}
    \centering
    \begin{subfigure}[b]{0.32\textwidth}
        \includegraphics[width=\textwidth]{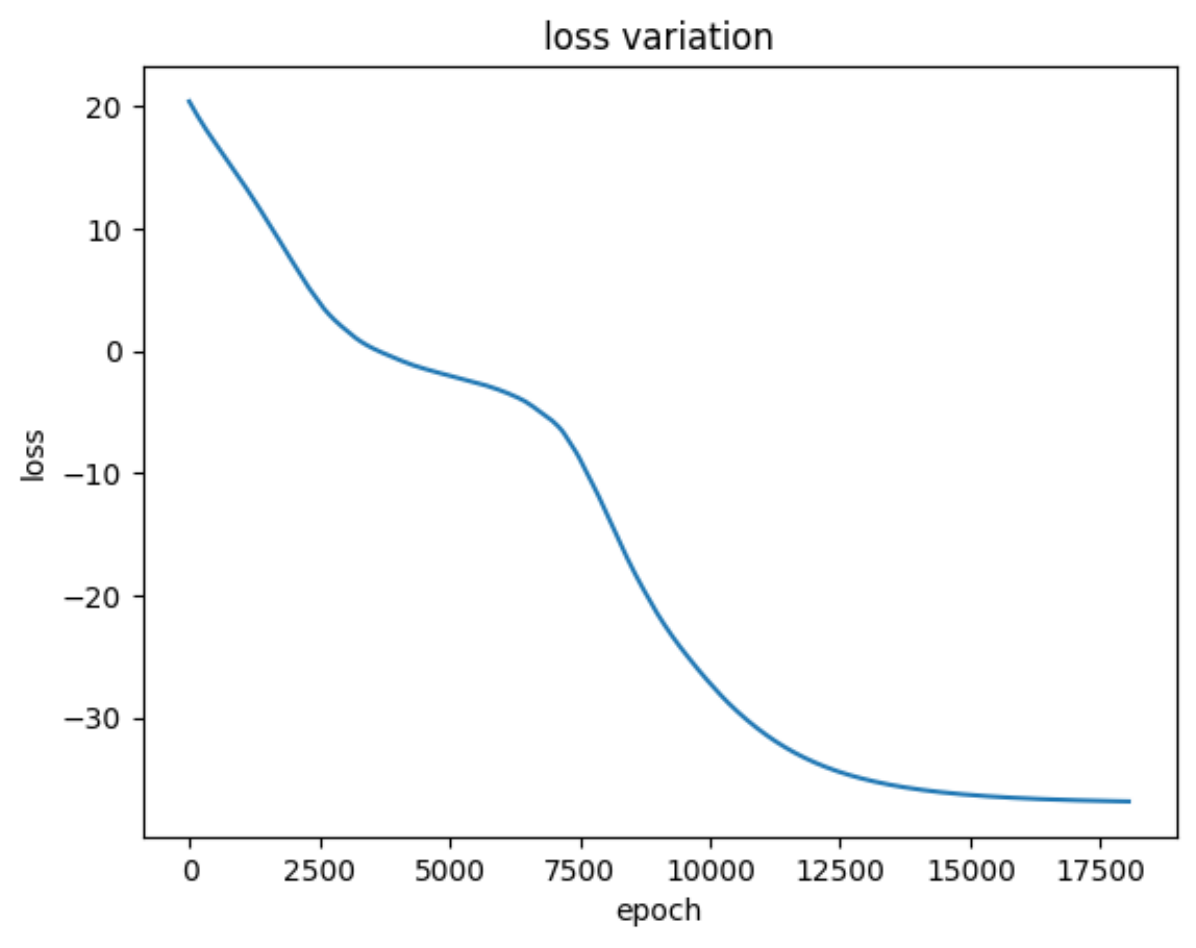}
    \caption{For the graph (50, 89)}
    \end{subfigure}
     \begin{subfigure}[b]{0.32\textwidth}
        \includegraphics[width=\textwidth]{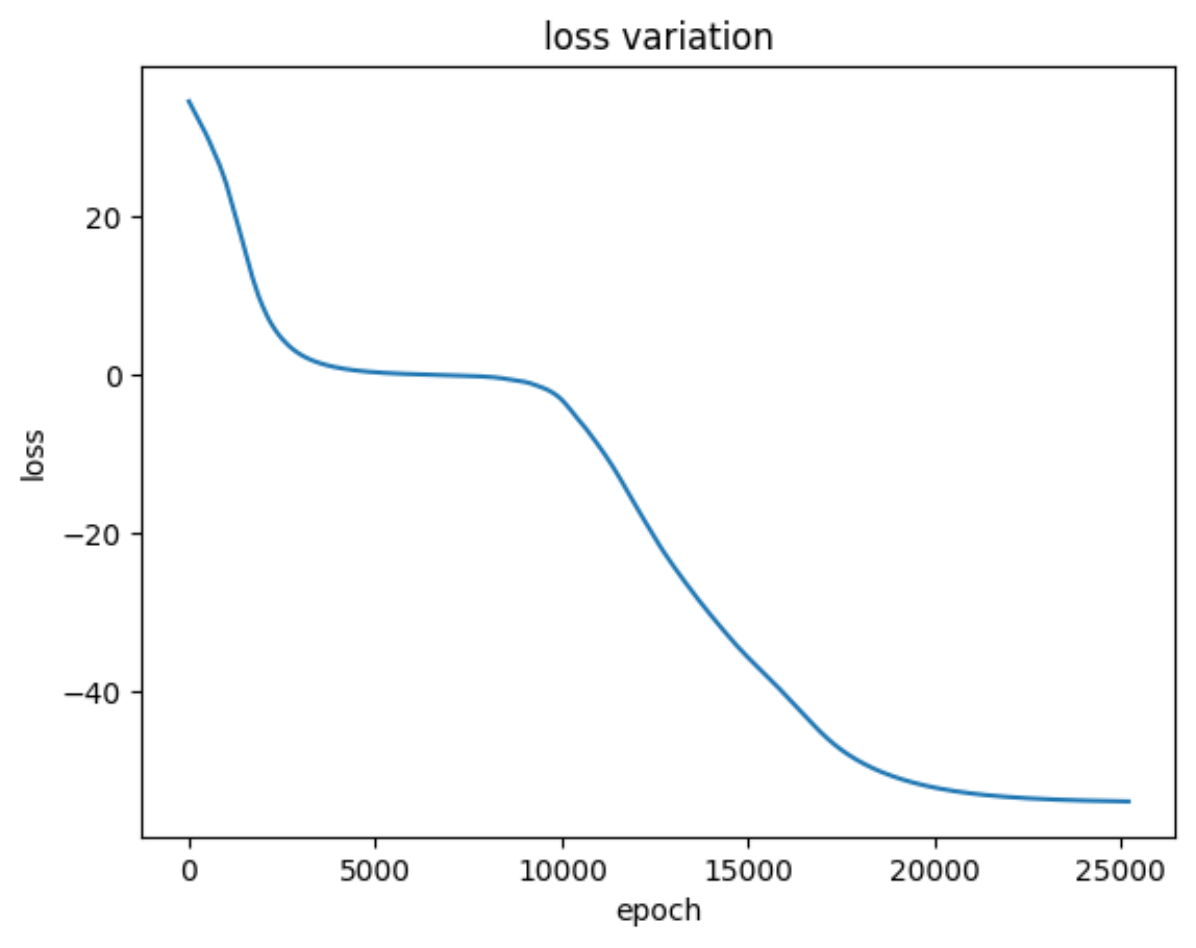}
    \caption{For the graph (50, 139)}
    \end{subfigure}
     \begin{subfigure}[b]{0.32\textwidth}
        \includegraphics[width=\textwidth]{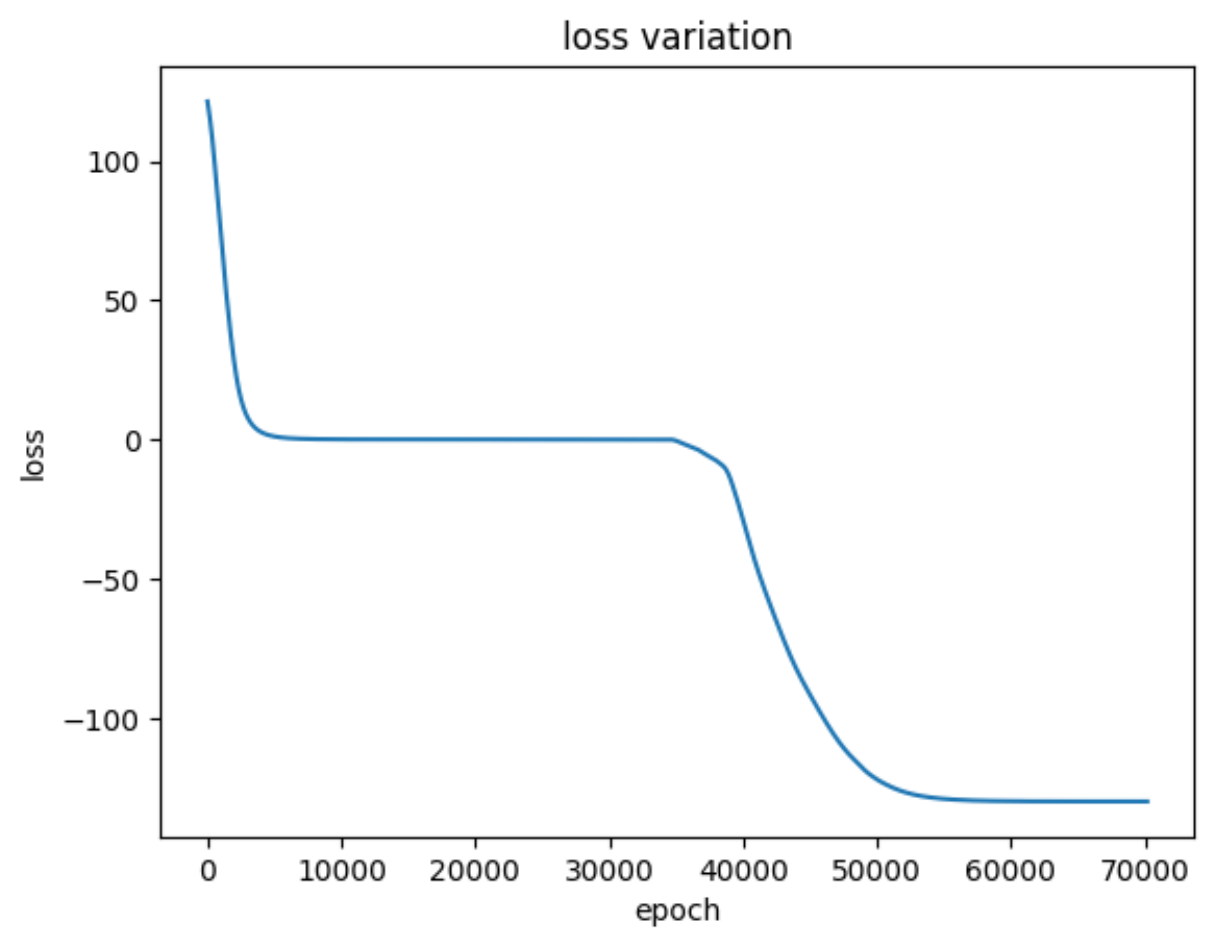}
    \caption{For the graph (50, 499)}
    \end{subfigure}
    \caption{Loss variation curve for PI-GNN}
    \label{fig:loss_pi_gnn}
\end{figure}

In this section, we mainly highlight the convergence status or training stability of the experimented architectures based on our set hyperparameters. We centralize our discussion based on three graphs of $50$ nodes with $89$, $139$, and $499$ edges, respectively for all the architectures to understand the behavior transition from sparser to denser graphs. In Fig. \ref{fig:50_89} we present the reward variations for PI-GNN, $GRL_{QUBO}$ and MCTS-GNN, respectively for a graph of 50 nodes with 89 edges. In a similar manner, we present the reward variations for a graph of 50 nodes with 139 edges and a graph of 50 nodes with 499 edges for all the architectures in Fig. \ref{fig:50_139} and \ref{fig:50_499}, respectively. To specifically understand the loss convergence status of PI-GNN, we present Fig. \ref{fig:loss_pi_gnn} for the three 50 node graphs with 89, 139, and 499 edges, respectively. Here, in the captions (50, 89) means a graph having 50 nodes and 89 edges.

The main observation that can be drawn from each is that the convergence behavior or almost linear variation is quite visible where our training has stopped for all the architectures. This supports the stability quality of the set hyperparameters, especially the patience values. Also, we do not provide all the other graphs' training convergence chart in this study, because they exhibit a similar pattern. All of our experiments are reproducible and can be regenerated by importing our official code repository.

\subsection{Complexity Analysis}
In this section, we present the time complexity of each of the experimented architectures or more specifically discuss the computation-intensive segments for each.

PI-GNN is the most scalable and runtime efficient compared to $GRL_{QUBO}$ and MCTS-GNN due to the novelties of GNN architecture. In simpler terms, to process a graph having $n$ nodes to generate a $d$ dimensional node feature vectors through $L$ layers of GNN, the \textcolor{black}{worst case} complexity is $\mathcal{O}(Lnd^{2})$\footnote{Here our assumption is that, the graph is sparse so, the number of edges $E=\mathcal{O}(n)$}. $GRL_{QUBO}$ also uses a GNN-variation (GAT) to encode the input graph. After that, it applies greedy selection at each iteration to pick a node with the highest attention value reported by the decoder. The additional complexity of this selection portion \textcolor{black}{takes a complexity of $\mathcal{O}(nlogn)$}. MCTS-GNN also uses a single GCN architecture in the rollout phases to build the search tree. So, a similar complexity of GNN is automatically added to the overall complexity. It also takes multiple attempts to train the same GNN again by varying inputs through manual perturbation of labelings. So, this incurs additional costs also. But our experiments suggest that only a very few times, GNN is trained exhaustively, and the other times, it generally runs for a very small number of epochs (less than $1000$ in our experiments) and reaches the early stopping criterion. So, in summary, PI-GNN is the most scalable. The complexity of $GRL_{QUBO}$ increases with the graphs becoming larger and the complexity of MCTS-GNN increases with the number of iterations to run to expand the search tree. \textcolor{black}{However, with this additional cost, we improve the quality of the final result by increasing the number of satisfied constraints compared to PI-GNN.}

Apart from these theoretical aspects, other constraints, e.g., step size or learning rate are also quite important and play a crucial role in converging the objective loss functions. If it takes a good time for the convergence, then the overall training time increases also. Based on our observations, generally, we found MCTS-GNN to take the most amount of time due to the expansion of the search tree. $GRL_{QUBO}$ takes way less time than MCTS-GNN and provides a competitive performance against PI-GNN - where PI-GNN takes a longer number of epochs to be trained, $GRL_{QUBO}$ takes way less number of epochs where there lies a significant amount of processing in each epoch by selecting the nodes in a greedy manner. PI-GNN shows worse performance than $GRL_{QUBO}$, especially in denser graphs where it might need a good amount of time to be converged. In Table \ref{tab:time}, we present some results, each value is denoted in seconds. But, a point to be mentioned is that, based on loss convergence status, in multiple trials, the runtime can vary significantly.

\begin{table}[]
    \centering
    \begin{tabular}{c|c|c|c|c}
       Node  & Edge & PI-GNN & $GRL_{QUBO}$ & MCTS-GNN\\
       \hline
       50 & 89 & 698 & 211 & 1066 \\
       50 & 139 & 691 & 184 & 835 \\
       50 & 499 & 3074 & 439 & 1834\\
       \hline
       100 & 199 & 317 & 347 & 5147\\
       100 & 799  & 2668 & 602 & 8479\\
    \hline
    300 & 399  & 398 & 427 & 1027\\
    300 & 899  & 667 & 862 & 5789\\
    300 & 1299  & 1837 & 1789 & 9035\\
    \hline
    500 & 799  & 478 & 679 & 1507\\
    500 & 1499  & 1478 & 1025 & 6478 \\
    500 & 5499  & 2478 & 2247 & 8798 \\
    \hline
    700 & 1199  &  932 & 725 & 2017 \\
    700 & 1699  &  1027 & 879 & 4789 \\
    700 & 4699  & 2104 & 2578 & 9786 \\
    \hline
    1000 & 1299  & 1265 & 1681 & 10510\\
    1000 & 3299  & 2333 & 1414 & 15146\\
    1000 & 5299  & 2017 & 2768 & 22147\\
    \hline
    3000 & 3499  & 1879 & 2147 & 8978 \\
    3000 & 4499  & 2998 & 2378 & 13147 \\
    3000 & 6999  & 5014 & 4789 & 24789 \\
    \hline
    \end{tabular}
    \caption{Training time in seconds for PI-GNN, $GRL_{QUBO}$, and MCTS-GNN for different graphs}
    \label{tab:time}
\end{table}

\section{Conclusion}
\textcolor{black}{In this study, we have extended the work of PI-GNN by improving its performance in terms of a number of satisfied constraints. We identify a crucial bottleneck issue observed \textcolor{black}{in graphs with higher densities} while applying PI-GNN during the transition of the loss values. We \textcolor{black}{proposed} a Fuzzy-stopping strategy in this regard to improve the quality of the solution. We also mentioned the issue regarding the absence of the node label's actual projection status in the QUBO-formulated Hamiltonian loss function and raised a concern that avoiding this factor might degrade the performance while applying the actual binary projection over the variables. In this regard, we proposed a Monty Carlo Tree Search with the GNN-based solution. This solution applies manual perturbation of node labels in the Hamiltonian loss function to guide a single GNN training while expanding the search tree. Furthermore, we also investigated the applicability of QUBO-formulated Hamiltonian in terms of reward function in RL setups. This contribution works as a bridge between RL-based solutions of CO problems with QUBO-based formulations while broadening the scope of applying actual node projection status during training. Our empirical result suggested that RL-based setups generally gave a better performance in terms of satisfied constraints than the PI-GNN setup with an additional incurring runtime or processing costs. So, our summarized observation can be included as PI-GNN is quite scalable and provides a moderate performance in terms of the number of satisfied constraints which can be improved by enforcing RL-based formulations. In the subsequent phase of our research, we intend to conduct a comprehensive investigation into enhanced reinforcement learning-based formulations with respect to combinatorial optimization challenges, taking into account factors such as the schematic construction of solutions, scalability potential, training paradigms, and graphical representation of problems, among others.}

\end{document}